\documentclass[journal]{IEEEtran}
\usepackage{cite}

\ifCLASSINFOpdf

\else

\fi

\usepackage{amsmath}

\usepackage{xcolor}
\usepackage{diagbox}
\usepackage{graphicx}
\usepackage{subfigure}
\usepackage{threeparttable}
\usepackage{booktabs}
\usepackage{array}
\usepackage{multirow}
\usepackage{microtype}
\usepackage{indentfirst}
\usepackage{tabularx}
\usepackage{stfloats}
\usepackage{array}
\usepackage{float}
\usepackage{url}
\usepackage{mathrsfs}
\usepackage{amsmath}
\usepackage{amsfonts,amssymb}
\usepackage{CJK}
\usepackage{amsthm,amsmath,amssymb}
\usepackage{bm}
\usepackage{subfigure}
\usepackage{makecell}

\begin{document}

\title{Adaptive Mixed-Scale Feature Fusion Network for Blind AI-Generated Image Quality Assessment}

\author{Tianwei Zhou, Songbai Tan, Wei Zhou, Yu Luo, Yuan-Gen Wang, and Guanghui Yue}

\markboth{}
{Shell \MakeLowercase{\textit{et al.}}: Bare Demo of IEEEtran.cls for IEEE Journals}

\maketitle

\begin{abstract}
With the increasing maturity of the text-to-image and image-to-image generative models, AI-generated images (AGIs) have shown great application potential in advertisement, entertainment, education, social media, etc. Although remarkable advancements have been achieved in generative models, very few efforts have been paid to design relevant quality assessment models. In this paper, we propose a novel blind image quality assessment (IQA) network, named AMFF-Net, for AGIs. AMFF-Net evaluates AGI quality from three dimensions, i.e., ``visual quality'', ``authenticity'', and ``consistency''. Specifically, inspired by the characteristics of the human visual system and motivated by the observation that ``visual quality'' and ``authenticity'' are characterized by both local and global aspects, AMFF-Net scales the image up and down and takes the scaled images and original-sized image as the inputs to obtain multi-scale features. After that, an Adaptive Feature Fusion (AFF) block is used to adaptively fuse the multi-scale features with learnable weights. In addition, considering the correlation between the image and prompt, AMFF-Net compares the semantic features from text encoder and image encoder to evaluate the text-to-image alignment. We carry out extensive experiments on three AGI quality assessment databases, and the experimental results show that our AMFF-Net obtains better performance than nine state-of-the-art blind IQA methods. The results of ablation experiments further demonstrate the effectiveness of the proposed multi-scale input strategy and AFF block.
\end{abstract}

\IEEEpeerreviewmaketitle

\section{Introduction}
\label{sec:intro}
\IEEEPARstart{W}{ith} the advent of the Web3.0 era\cite{gan2023web}, artificial intelligence-generated Content (AIGC) is quietly leading a profound change, reshaping and even subverting the production and consumption mode of digital content. As a major branch of AIGC, AI-Generated Images (AGIs) have shown great application potential in various aspects of human life. The creation of AGIs involves inputting text prompts into the text-to-image generative model or inputting an image into the image-to-image generative model to facilitate the generation process \cite{frolov2021adversarial}. However, due to technical limitations, large quality variance exists among different AGIs, requiring manual selection before use, which greatly limits the development of generative techniques. Therefore, to improve production efficiency, it is of great significance to design an automatic AGI image quality assessment (IQA) method \cite{li2023agiqa,goring2023appeal}.

Recently, deep neural networks (DNNs), especially Convolutional Neural Networks (CNNs) and Transformers, have been widely employed in IQA tasks due to their outstanding capabilities in feature extraction and fitting \cite{lan2023multilevel,yue2024subjective,1037556601,yue2022improving,lang2023full,10337739}. Current works mainly focus on natural scene images (NSIs) that include synthesized distortions or authentic distortions. Most existing DNN-based IQA methods started with automatically mining quality-aware features through newly designed shallow network architectures \cite{kim2017deep} or classical network architectures used in image classification tasks \cite{hu2023blind}. To strengthen the representative capabilities of DNN features, designers also focused on the development of more comprehensive feature extraction or fusion approaches \cite{bosse2017deep,10440553,ma2017end,10464346}. Later, for more accurate assessment results, some strategies are used in view of the distortion knowledge during the network design, such as setting auxiliary tasks for naturalness evaluation \cite{song2022knowledge}, combing local and global features \cite{golestaneh2022no}, generating pseudo reference \cite{pan2022vcrnet}, integrating multi-level features \cite{lan2023multilevel}, etc. In addition, some works also considered the characteristics of the human visual system (HVS) in the process of network design, e.g., establishing perception rule \cite{su2020blindly}, including visual saliency prediction as the auxiliary task \cite{yang2019sgdnet}, modeling the attention and contrast sensitivity mechanisms \cite{you2022attention}. Recently, some works also proposed to utilize rank learning \cite{ou2021novel}, semi-supervised learning \cite{yue2022semi}, and contrastive learning \cite{yang2023multi} for robust feature representation, striving for more accurate assessment results.

\begin{figure}[!htpb]
  \centering
  \includegraphics[width=0.92\columnwidth]{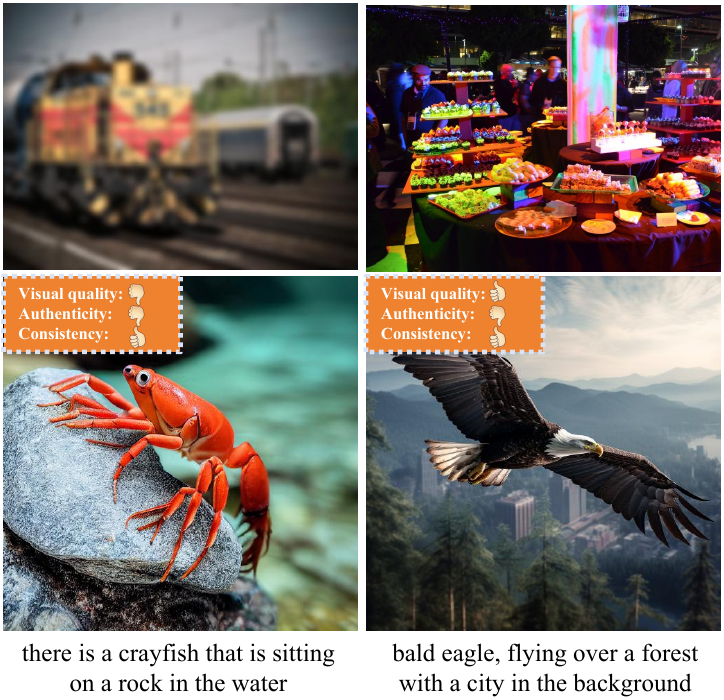}\\
  \caption{Comparisons between NSIs (the first row, selected from KADID-10k \cite{lin2019kadid} and KonIQ-10k \cite{lin2018koniq}) and AGIs (the second row, selected from PKU-I2IQA\cite{yuan2023pku}). The quality score of NSIs is mainly rated in the dimension of visual experience affected by distortions, while the quality scores of AGIs are rated in the dimensions of visual quality affected by distortions, authenticity affected by realness degree to reality, and consistency affected by the alignment between image content and textual labels.}
  \label{Fig.1}
\end{figure}

However, unlike camera-captured NSIs, AGIs are directly generated by AI generative models. Fig. \ref{Fig.1} shows a simple comparison between NSIs and AGIs. Generally, NSIs usually suffer from distortions (e.g., compression, blurriness, noise, etc.), and the perceptual quality is rated from what is the level of visual experience. By contrast, the quality definition and representation of AGIs are different and usually rated in a multi-dimensional perspective, typically in the form of visual quality, authenticity, and consistency  \cite{frolov2021adversarial}. For an AGI, visual quality is similar to the perceptual quality of NSIs that is rated by analyzing the visual experience influenced by distortions, such as compression and color artifacts. Authenticity measures the realness degree to reality. Consistency is gauged by the alignment between image content and textual labels. Generally, the rated scores of these three dimensions are usually different. For instance, in the last subfigure of Fig. \ref{Fig.1}, the generated bald eagle is clear, without obvious distortions, and the image content conforms to the text description. But the image doesn't give enough details about the bald eagle with stiff hair and coarse structures, making it easy to identify as fake. Therefore, the rating scores of visual quality and consistency of this image are high, while the rating score of authenticity is low. In practical applications, apart from the visual quality affected by the distortion and authenticity affected by the artistic expression, users also concern about how well the generated AGI matches the task, i.e., content consistency. Since traditional NSI oriented IQA methods can only evaluate the image quality in the dimension of visual experience, they are not suitable for the quality assessment task of AGIs. Therefore, it is necessary to design a comprehensive quality evaluation method to form more detailed understanding of AGIs.

As a new topic, the research on AGI quality assessment is still in infancy, with very limited progress. A widely used strategy for evaluating the quality of AGIs is calculating the distance between the NSIs groups and the AGIs groups \cite{heusel2017gans,binkowski2018demystifying}. However, such a strategy must evaluate a group of images, which is unsuitable for the case of only one image. Therefore, more advanced methods should be specifically proposed. Following the general steps of IQA tasks, the quality assessment databases were first reported based on subjective experiments to promote the development of objective IQA methods \cite{li2023agiqa,wang2023aigciqa2023,yuan2023pku}. These reported databases contain text prompts and multi-dimensional quality labels. Alongside these databases, some mainstream NSI oriented IQA methods were tested by respectively retraining them multiple times, each with a specific dimension of quality labels as the ground truth. As a result, most existing methods perform poorly, especially on the task of evaluating content consistency, as they only take the image as the input and cannot measure the mismatch degree between the text prompt and generated image. Recently, some works proposed to process the text prompt and image separately with different encoders and concatenate the extracted features from two encoders to generate the quality score \cite{yuan2024tier}. However, directly concatenating features cannot effectively measure the semantic difference between the text prompt and image. For detailed alignment measurement between the text prompt and image, some researches tried to segment the text prompt into multiple morphemes, cut the image into multiple patches, and compute the alignment scores between morphemes and sub-images one by one \cite{li2023agiqa}. However, image cutting is highly depended on the designer's experience and how to build the correspondence between morphemes and sub-images is unclear. To sum up, current studies mainly stay at benchmarking these databases with some mainstream IQA methods, lacking in-depth research on designing AGI quality assessment methods. 

To move this field forward, this paper proposes a novel Adaptive Mixed-Scale Feature Fusion Network (AMFF-Net) for blind AGI quality assessment. Specifically, AMFF-Net adopts a multi-task framework and evaluates the quality of AGIs from three dimensions: distortion, authenticity, and content consistency. Considering that the subjective evaluation result of an image varies when the distance of the image plane from the observer changing, AMFF-Net scales the AGI up and down and feeds the scaled images and the original-sized image into the image encoder of the pre-trained Contrastive Language-Image Pre-Training (CLIP) model \cite{radford2021learning}. After that, the extracted multi-scale features are adaptively fused by an Adaptive Feature Fusion (AFF) block. The fused features contain information at different scales of the image, thereby being more effective for characterizing the distortion and authenticity of AGIs. For content consistency prediction, AMFF-Net uses the text encoder in the pre-trained CLIP to encode the text prompt and computes the similarity between the obtained textual features and the fused multi-scale features. The main contributions of this paper can be summarized as follows:
\begin{itemize}
  \item
  A novel blind IQA method is proposed to comprehensively evaluate the quality of AGIs. Different from existing works that only measure the ``visual quality'' of an image, our method evaluates an AGI from the perspectives of ``visual quality'', ``authenticity'', and ``consistency''. %
  \item
  Given that both local and global information should be considered during subjective rating of visual quality and authenticity, we propose to utilize a multi-scale input strategy to help the network capture image details at different levels of granularity.
  \item
  An AFF block is proposed to fuse multi-scale features. Different from current works that directly concatenate or add multi-scale features, the proposed block adaptively calculates the weights for different features, reducing the risk of information masking caused by concatenation and addition. 
  \item
  Extensive experiments on three AGI quality assessment databases show that AMFF-Net achieves superior results compared to nine state-of-the-art blind IQA methods. Ablation experiments further demonstrate the effectiveness of the multi-scale input strategy and the AFF block.
\end{itemize}

\section{Related work}
\label{sec:format}
\subsection{DNN-based Blind IQA Methods for NSIs}
In the past decades, DNN-based blind IQA methods have attracted increasing attention from scholars, obtaining superior performance over traditional handcrafted feature-based methods on multiple tasks \cite{10464346,zhou2023perception}. In earlier studies, shallow CNNs were constructed to build the mapping between image patches and quality scores \cite{kim2017deep}. For instance, Bosse \emph{et al}.\cite{bosse2017deep} proposed a CNN-based blind IQA network, which consists of only ten convolutional layers and five pooling layers for feature extraction and two fully connected layers for quality regression. Since the constructed shallow networks are usually with small input size, they have limited ability to characterize global distortions \cite{yue2019blind}. For more accurate predictions, latter works mainly utilized the pre-trained CNNs, Transformers, or the hybrid for the image classification tasks as the backbone of the IQA network. These networks usually have a relatively large input size, which is conducive to extract global information. Su \emph{et al}. \cite{su2020blindly} utilized the pre-trained ResNet \cite{he2016deep} as the backbone to extract local and global features and fused these features with a hyper network. Golestaneh \emph{et al}. \cite{golestaneh2022no} extracted local information of the image via a pre-trained CNN and modeled them as a sequential input to a Transformer block for the non-local representation. Ke \emph{et al}. \cite{ke2021musiq} proposed a Transformer-based blind IQA network, in which the native resolution images with varying sizes are processed for a multi-scale image representation. 

Generally, the aforementioned methods are purely data-driven and do not fully consider the characteristics of distortions or HVS, leaving much room for performance improvement. Recently, some researches proposed to build multi-task IQA frameworks for better performance, in which the auxiliary task is related to distortion understanding or HVS-inspired predictions. For instance, given that different distortions have diverse impacts on the perceptual quality, Wu \emph{et al}. \cite{wu2014no} designed a multi-task IQA network that takes image distortion type recognition as the auxiliary task. Yang \emph{et al}. \cite{yang2019sgdnet} set the visual saliency prediction as the auxiliary task in view of that different regions in an image receive non-uniform visual attention as they exhibit different qualities. Considering that synthesized distortions can be characterized and quantified by natural scene statistics (NSS), Yan \emph{et al}. \cite{yan2019naturalness} set the NSS feature prediction as the auxiliary task to help the main task, i.e., quality prediction, learn a better mapping between the image and its quality score. Song \emph{et al}. \cite{song2022knowledge} proposed a knowledge-guided blind IQA framework by integrating domain knowledge from NSS and HVS. However, simply utilizing multi-task learning may have limited feature representations when dealing with images with complex distortions. Sun \emph{et al}. \cite{sun2023blind} proposed a staircase structure to hierarchically fuse low-level and high-level features for better feature representation.

\subsection{DNN-based Blind IQA Methods for AGIs}
Compared to NSIs, the quality research of AGIs is still in its infancy, with only a few explorations. For a long time, researchers mainly utilized the \emph{Inception Scor}e proposed by Salimans \emph{et al}. \cite{salimans2016improved} for AGI quality evaluation. Considering the difference between NSIs and AGIs, the \emph{Frechet Inception Distance} \cite{heusel2017gans} and \emph{Kernel Inception Distance} \cite{binkowski2018demystifying} were later designed to measure the quality of AGIs by calculating the distance between the AGI groups and the NSI groups. Since these metrics only evaluate the quality of AGIs in a single dimension and are not suitable for evaluating a single AGI, more advanced metrics are highly required. Zhang \emph{et al}. \cite{zhang2023perceptual} made one of the pioneering discussions on evaluating AGI quality in a multi-dimensional manner and suggested that the quality of AGIs should be measured in the aspects of technical issues, artificial intelligence, unnaturalness, degree of difference, and aesthetics. Unfortunately, this work does not propose an AGI quality evaluation algorithm, and the strategy of how to incorporate AGI distortion representation into the evaluation algorithm is not clear. More recently, some efforts have been paid to propose specific IQA methods for AGIs by drawing inspirations from NSI-oriented IQA methods. For instance, Yuan \emph{et al}. \cite{yuan2023pscr} took the AGI as the input of the IQA network and compared the differences among various images for a better feature representation using a contrastive regression framework. Later, they \cite{yuan2024tier} also introduced a text-image encoder-based regression framework that respectively processes the text prompts and generated images with a text encoder and an image encoder, and concatenates the extracted features for quality prediction. To evaluate the consistency between the text prompt and generated image, some current works leverage the strong reasoning abilities of large Language models (LLMs) for evaluation \cite{kirstain2024pick, lu2024llmscore, wu2023human}. For example, LLMScore\cite{lu2024llmscore} generates quality scores with multi-granularity compositionality, which transforms the image into image-level and object-level visual descriptions, leveraging LLMs to evaluate text-to-image models. However, these LLMs based methods have a large number of parameters and require abundant labeled AGIs databases for training. This motivates to design simpler methods. Li \emph{et al}. \cite{li2023agiqa} proposed a simply StairReward alignment model, which segments the prompt into morphemes and divides the image into stairs to predict the final score through their one-to-one alignment.

Although these AGI quality assessment methods perform relatively superior performance over traditional NSI-oriented IQA methods, they are still not fully qualified for practical applications due to the following limitations. First, most methods only predict the visual experience of an image. In practice, before use, users not only comprehensively check the AGI from the visual experience and authenticity, but also concern about how well the generated image matches the task, i.e., content consistency. Second, most methods only take the AGI as the input, which is insufficient to measure the text-to-image consistency. Although the measurement of alignment degree between text morphemes and image stairs can reflect the consistency to some extent, the prompt segmentation and image cutting are highly dependent on the designer's experience, limiting the generalization ability of such methods. Third, current methods ignore the correlation and interaction between features from different modalities, usually resulting in unsatisfactory performance.

\subsection{CLIP-based IQA Methods}
In 2021, Radford \emph{et al}. \cite{radford2021learning} trained and released the CLIP model based on 400 million picture-text pairs to enable zero-shot prediction. Since then, many researchers have proposed CLIP-based models for many visual tasks, including image classification \cite{abdelfattah2023cdul}, object detection \cite{teng2021global}, and image retrieval \cite{sain2023clip}. Thanks to the robust capabilities of semantic extraction, CLIP has recently been applied to the IQA tasks. Wang \emph{et al}.\cite{wang2023exploring} explored the rich prior knowledge and visual-perception in CLIP, and evaluated the image quality through prompt engineering. Pan \emph{et al}.\cite{pan2023quality} introduced a two-stage IQA model, in which a CLIP-based text encoder is used for quality-aware feature extraction. Miyata \cite{miyata2023interpretable} introduced a CLIP-based IQA model that can identify both the perceived quality rating of the image and the reason on which the rating is based. Zhang \emph{et al}. \cite{zhang2023advancing} utilized the CLIP model to measure semantic affinity in the digital human quality assessment task. Although increasing attempts made in designing CLIP-based IQA methods for NSIs, very few works have been reported for AGIs. 

\section{Proposed Method}
\label{sec:format}
\subsection{Motivation}
In this paper, we propose a simple yet effective multi-task blind IQA network, named AMFF-Net, for AGIs. The design motivation behind our AMFF-Net are as follows: $1)$ Inspired by the fact that the perceivability of image details is highly related to the distance of the image plane from the observer, AMFF-Net scales the AGI up and down, and encodes the scaled images and original-size image to capture image details at different levels of granularity, closer to the subtle and holistic nature of human visual perception. 2) With the multi-scale image representations, an adaptive feature fusion block is used to adaptively incorporate image details at different resolutions, contributing to accurate quality and authenticity predictions. 3) AMFF-Net evaluates the AGI from multi-dimensional perspectives, i.e., the visual quality, authenticity, and content consistency, targeting at helping the users understand the quality of AGIs more comprehensively than current methods that only provide one-dimensional prediction. Also, computing the similarity between textual features and image features requiring no designer's experience is a good choice to evaluate the consistency between the prompt and generated image.

\subsection{Overall Architecture}
\label{ssec:subhead}
\begin{figure}[!htpb]
  \centering
  \includegraphics[width=\columnwidth]{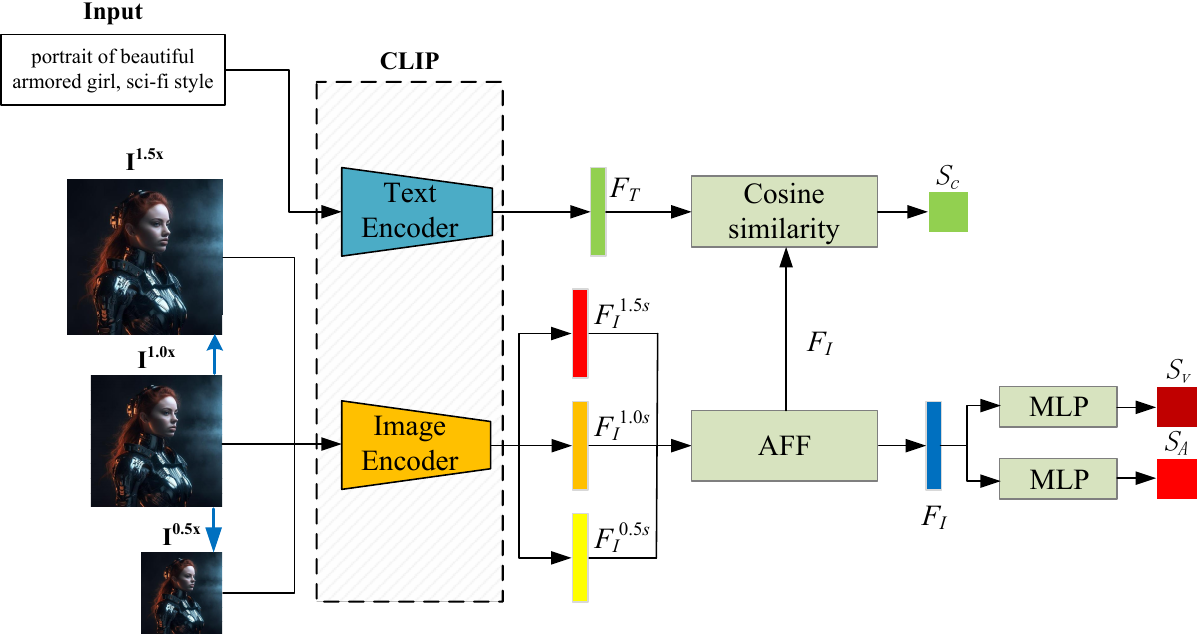}\\
  \caption{Overview of the proposed AMFF-Net. It takes both the text prompt and three scaled AGIs as the inputs and outputs the consistency score $S_C$, visual quality score $S_V$, and authenticity score $S_A$. Here, the image and text encoders are selected from the pre-trained CLIP model. AFF and MLP denote the adaptive feature fusion block and multi-layer perception, respectively.}\label{Fig.2}
\end{figure}

Fig. \ref{Fig.2} illustrates the overall architecture of the proposed AMFF-Net. AMFF-Net takes the text prompt and multi-scale AGIs as the inputs and outputs multi-attribute quality scores, including content consistency $S_C$, visual quality $S_V$, and authenticity $S_A$:
\begin{equation}\label{Eq.1}
  (S_C, S_V, S_A) = \mathcal{F}_\theta(T, I^{1.5\times}, I^{1.0\times}, I^{0.5\times}),
\end{equation}
where $\mathcal{F}_\theta(\cdot,\cdot,\cdot,\cdot)$ denotes the mapping between inputs and outputs, and $T$ is the text prompt. $I^{1.5\times}$, $I^{1.0\times}$, and $I^{0.5\times}$ are the scaled AGIs with $1.5\times$, $1.0\times$, and $0.5\times$ resolutions of the original image.

Specifically, considering that both local and global details affect the subjective ratings of visual quality and authenticity, AMFF-Net inputs the scaled AGIs, i.e., $I^{1.5\times}$, $I^{1.0\times}$, and $I^{0.5\times}$, into an image encoder of the pre-trained CLIP model \cite{radford2021learning} to obtain multi-scale semantic representations, denoted as $F_I^{1.5s}$ $\in$ $\mathbb{R}^{1\times1024}$, $F_I^{1.0s}$ $\in$ $\mathbb{R}^{1\times1024}$, and $F_I^{0.5s}$ $\in$ $\mathbb{R}^{1\times1024}$. In this study, we select ResNet50 \cite{he2016deep} as the image encoder as it has been widely validated to be effective for IQA tasks. In the default settings of CLIP, the image encoder can only accept inputs with the size of $224\times224$ due to the presence of positional embedding. In this study, we add some operations on the last layer of ResNet50 to make it adaptive to inputs with different sizes. Specifically, for the input with larger size than $224\times224$, we down-sample the feature of the ResNet50's last layer using adaptive maximum pooling. For the input with smaller size than $224\times224$, we up-sample the feature of the ResNet50's last layer using the bilinear interpolation. Both the down-sampling and up-sampling operations aim to make the feature size of the ResNet50's last layer be (2048, 7, 7) to match the positional embedding. The obtained semantic features of different scaled images are then fused by an AFF block to form the merged feature $F_I$ $\in$ $\mathbb{R}^{1\times1024}$. Subsequently, $F_I$ is fed into two parallel Multi-layer Perceptions (MLPs), which consists of two fully connected layers, to separately predict the scores of visual quality and authenticity:
\begin{equation}\label{Eq.2}
\begin{cases}
 S_V=\mathcal{M}_{\vartheta1}(F_I), \\
 S_A=\mathcal{M}_{\vartheta2}(F_I),
\end{cases}
\end{equation}
where $\mathcal{M}_{\vartheta1}(\cdot)$ and $\mathcal{M}_{\vartheta2}(\cdot)$ are the MLP operations for visual quality score prediction and authenticity score prediction, respectively. The number of neural nodes in the MLP is $\{1024, 256, 1\}$. For content consistency score prediction, we first utilize the Transformer-based text encoder \cite{vaswani2017attention} of the pre-trained CLIP model to encode the text prompt, and subsequently compute the cosine similarity between the extracted textural feature $F_T$ $\in$ $\mathbb{R}^{1\times1024}$ with the merged image feature $F_I$ to predict the content consistency score:
\begin{equation}\label{Eq.3}
  S_C = \frac{F_I\odot\left(F_T\right)^T}{\left\|F_I\right\|_2\left\|F_T\right\|_2},
\end{equation}
where $\odot$ denotes the matrix-multiplication, and $(\cdot)^T$ stands for the matrix transpose operation.

Overall, AMFF-Net can extract multi-scale representations and adaptively aggregate them, providing rich semantic information for accurate visual quality and authenticity predictions. In addition, AMFF-Net considers the interaction between the prompt and image, which is conducive to accurate content consistency prediction. 

\subsection{Adaptive Feature Fusion Block}
\label{ssec:subhead}
By feeding three scaled AGIs into the image encoder, we can obtain multi-scale semantic representations, i.e., $F_I^{1.5s}$, $F_I^{1.0s}$, and $F_I^{0.5s}$. How to fuse these features is our next focus. In previous works, one widely used method is directly concatenating or adding these features. However, such a method is easy to cause information masking at different scales, which is unsuitable for the IQA task. In this study, we propose a novel feature fusion block, named AFF, to adaptively fuse these features. 

\begin{figure}[!htpb]
  \centering
  \includegraphics[width=0.95\columnwidth]{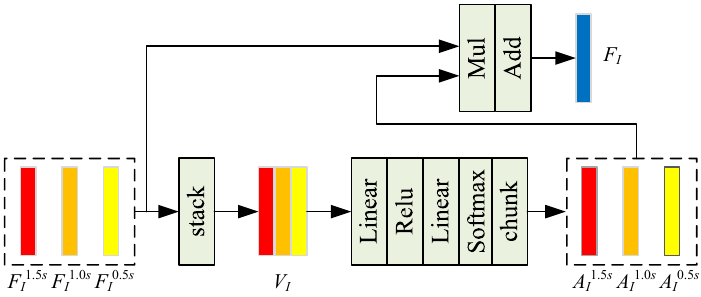}\\
  \caption{Architecture presentation of the proposed AFF block.}\label{Fig.3}
\end{figure}

Fig. \ref{Fig.3} presents the architecture of the proposed AFF block. First, these multi-scale features are stacked together, resulting in $V_I$ $\in$ $\mathbb{R}^{3\times1024}$. Then, $V_I$ is processed by two linear layers and a Softmax function. Between two linear layers, a ReLU activation function is embedded. Next, the resultant feature is processed by a chunk operation, resulting in three distinct features, namely $A_I^{1.5s}$ $\in$ $\mathbb{R}^{1\times1024}$, $A_I^{1.0s}$ $\in$ $\mathbb{R}^{1\times1024}$, and $A_I^{0.5s}$ $\in$ $\mathbb{R}^{1\times1024}$:
\begin{equation}\label{Eq.4}
  V_m=\Psi(\mathcal{U}(F_I^{0.5s},F_I^{1.0s},F_I^{1.5s})),
\end{equation}
where $\Psi\left(\cdot\right)$ indicates the weights generation process, including the stack operation $\mathcal{U}$, linear layers, and a Softmax function. $A_I^{1.5}$, $A_I^{1.0}$, and $A_I^{0.5}$ represent the weights associated with the three image scales. With multi-scale semantic features and their scale-specific weights, we can obtain the fused feature $F_{I}$ $\in$ $\mathbb{R}^{1\times1024}$ through the element-multiplication and element-addition operations:
\begin{equation}\label{Eq.5}
  F_{I}=A_{I}^{1.5s}\cdot F_{I}^{1.5s}+A_{I}^{1.0s}\cdot F_{I}^{1.0s}+A_{I}^{0.5s}\cdot F_{I}^{0.5s}.
\end{equation}
This above process ensures an adaptive integration of multi-scale semantic features, boosting the feature representation. 

\subsection{Loss Function}
\label{ssec:subhead}
As shown in Fig. \ref{Fig.1}, our AMFF-Net evaluates the quality in three dimensions. The overall loss function $\mathcal{L}$ for training AMFF-Net is a linear combination of three components:
\begin{equation}\label{Eq.6}
\mathcal{L}=\mathcal{L}_C+\mathcal{L}_V+\mathcal{L}_A
\end{equation}
where $\mathcal{L}_C$, $\mathcal{L}_V$, and $\mathcal{L}_A$ are the loss functions respectively used for three tasks, i.e., content consistency score prediction, visual quality score prediction, and authenticity score prediction, during network training. For the task of consistency score prediction, we utilize the fidelity loss function \cite{tsai2007frank}:
\begin{equation}\label{Eq.7}
  \mathcal{L}_C=\frac1{N^2}\sum_{i,j\in N}\bigg(1-\sqrt{\mathcal{P}_{i,j}\widehat{\mathcal{P}}_{i,j}}-\sqrt{(1-\mathcal{P}_{i,j})(1-\widehat{\mathcal{P}}_{i,j})}\bigg),
\end{equation}
where $N$ is the image number in a mini-batch, and $\mathcal{P}_{i,j}$ is a binary function that compares the quality of the $i$-the and $j$-th images. If $Q_V^i\geq Q_V^j$, $\mathcal{P}_{i,j}=1$; otherwise, $\mathcal{P}_{i,j}=0$. Here, $Q_V$ denotes the ground truth value of content consistency. $\widehat{\mathcal{P}}_{i,j}$ computes the probability
of the $j$-th image predicted better than the $j$-th image using the Thurstone's model \cite{thurstone1994law}. The reason why we choose the fidelity loss is that it can preserve information granularity of the cosine similarity between image and text features, making it suitable for our image-text consistency score prediction task. For the tasks of visual quality and authenticity score prediction, the mean square error is used as the loss function to train the network:
\begin{equation}\label{Eq.8}
  \mathcal{L}_V=\frac{1}{N}\sum^{N}_{i=1}(Q_{V}^{n} -S_{V}^{n})^2,
\end{equation}
\begin{equation}\label{Eq.9}
  \mathcal{L}_A=\frac{1}{N}\sum^{N}_{i=1}(Q_{A}^{n} -S_{A}^{n})^2.
\end{equation}
In Eq. \eqref{Eq.8} and Eq. \eqref{Eq.9}, for a mini-batch with $N$ images, $Q_{V}^{n}$ and $Q_{A}^{n}$ are the ground truth values of the $n$-th image's visual quality and authenticity, respectively.

\section{Experiments}
\label{sec:pagestyle}

\subsection{Experiments Setup}
\label{ssec:4}
\begin{figure*}[t]%
  \centering
  \includegraphics[width=2.0\columnwidth]{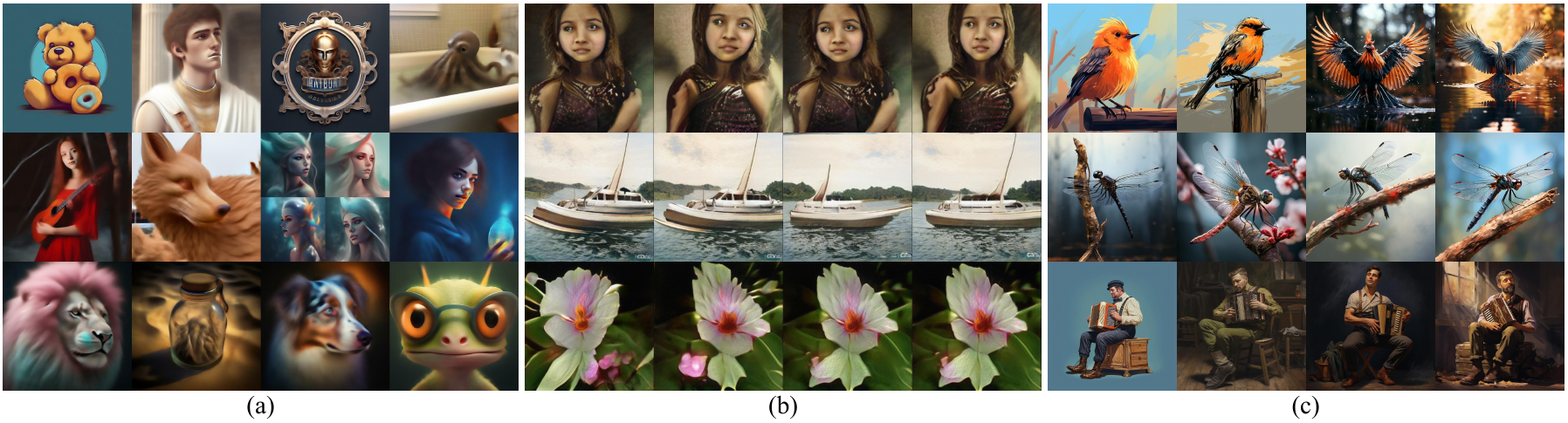}\\
  \caption{Images from three AGI quality assessment databases: (a) AGIQA-3K\cite{li2023agiqa}, (b) AIGCIQA2023\cite{wang2023aigciqa2023}, and (c) PKU-I2IQA\cite{yuan2023pku}.}\label{Fig.4}
\end{figure*}
\subsubsection{Databases}
\label{sssec:4.1}
In this study, we selected three public AGI quality assessment databases, including AGIQA-3K\cite{li2023agiqa}, AIGCIQA2023\cite{wang2023aigciqa2023}, and PKU-I2IQA\cite{yuan2023pku}, for evaluating and comparing different blind IQA methods. Brief descriptions of these databases are given below.
\begin{itemize}
  \item
\textbf{AGIQA-3K}: It has 2,982 AGIs generated by 6 Text-to-Image generative models, including GLIDE \cite{nichol2021glide}, Stable Diffusion V1.5 \cite{rombach2022high}, Stable Diffusion XL2.2 \cite{rombach2022high}, AttnGAN \cite{xu2018attngan}, Midjourney \cite{borji2022generated}, and DALLE2 \cite{ramesh2022hierarchical}. It also provides 300 text prompts of different scenes and styles, and contains score labels of visual quality and consistency.
  \item
\textbf{AIGCIQA2023}: It comprises 2,400 AGIs generated by six cutting-edge Text-to-Image generative models with 100 text prompts, including GLIDE \cite{nichol2021glide}, Lafite \cite{zhou2022towards}, Stable Diffusion \cite{rombach2022high}, DALLE \cite{ramesh2022hierarchical}, Unidiffusion \cite{bao2023one}, and Controlnet\cite{zhang2023adding}. For each image, it provides score labels of visual quality, authenticity, and consistency.
  \item
\textbf{PKU-I2IQA}: It consists of 1,600 AGIs generated by two Image-to-Image models (Midjourney \cite{borji2022generated} and Stable Diffusion V1.5 \cite{rombach2022high}). It contains score labels of visual quality, authenticity, and consistency, and also provides 200 text prompts of different scenes and styles.
\end{itemize}

Fig. \ref{Fig.4} shows some examples from the three AGI quality assessment databases. Following previous works, we randomly divide the AGIQA-3K into training set and test set in a ratio of 8:2. For AIGCIQA2023 and PKU-I2IQA, we conduct a 3:1 train-test split on the images generated by each generative model. The resolution of images in three databases is $512\times512$, and we resize each image into the size of $224\times224$ due to the limited computational source. 

\subsubsection{Evaluation Metrics}
\label{sssec:subsubhead}
This study selects three widely used evaluation metrics in the IQA field to present quantitative results, including Spearman rank-order correlation coefficient (SRCC), Pearson linear correlation coefficient (PLCC), and Kendall rank correlation coefficient (KRCC). In general, higher values (the maximum value is 1) of these metrics indicates better prediction performance. As suggested by VQEG \cite{antkowiak2000final}, a four-parameter logistic function is used before computing PLCC:
\begin{equation}\label{Eq.8}
  \widetilde{s}=\frac{\kappa_1 - \kappa_2}{1+exp(\kappa_4(s-\kappa_3))} + \kappa_2,
\end{equation}
where $\widetilde{s}$ is the mapped value of a predicted score $s$. In Eq. \eqref{Eq.8}, $\kappa_i$ ($i\in\{1,2,3,4\}$) can be obtained by comparing the predicted scores and their ground truths using the least square method.

\subsubsection{Implementation Details}
\label{sssec:subsubhead}
Our proposed AMFF-Net was implemented under the open source Pytorch repository. The server used in the experiments was powered by one NVIDIA Geforce GTX3090 GPU and two Intel XEON 6226R CPUs. During network training, the AdamW optimizer was used, and the batch size was set to 32. The model was trained end-to-end for 120 epochs in a two-stage manner. In the first stage, we froze the parameters of the text and image encoders in the CLIP and trained the remaining part of AMFF-Net in the first 20 epochs. The learning rates for AIGCIQA2023 and the other two databases were set to $1\times10^{-3}$ and $5\times10^{-4}$. In the second stage, we unfroze the text and image encoders and fine-tuned the whole network with a learning rate of $5\times10^{-6}$. At the 80-th epoch, the learning rate was adjusted to $5\times10^{-7}$. In addition, we set an early stop strategy, and the training process was stopped if the performance did not improve after 20 epochs.

\begin{table*}[!htpb]
  \centering
  \caption{Quantitative results comparison between different blind IQA methods on AGIQA-3K. For convenient viewing, we also present the average value of each evaluation metric on two tasks in the eighth to tenth columns of the table.}
  \label{table.1}
  \begin{threeparttable}
  \renewcommand{\arraystretch}{1.0}
  \begin{tabular}{c|ccccccccc|c|c}
    \toprule
    \multirow{3}{*}{Method} & \multicolumn{9}{c|}{AGIQA-3K\cite{li2023agiqa}} & \multirow{3}{*}{FLOPs} & \multirow{3}{*}{\#Params} \\ \cline{2-10}
     & \multicolumn{3}{c|}{Quality} & \multicolumn{3}{c|}{Consistency} & \multicolumn{3}{c|}{Avg.} &  &  \\ \cline{2-10}
     & SRCC & PLCC & \multicolumn{1}{c|}{KRCC} & SRCC & PLCC & \multicolumn{1}{c|}{KRCC} & SRCC & PLCC & KRCC &  &  \\ \hline
    ResNet50\cite{he2016deep}  & 0.8252 & 0.8863 & \multicolumn{1}{c|}{0.6396} & 0.6396 & 0.7878 & \multicolumn{1}{c|}{0.4645} & 0.7324 & 0.8370 & 0.5521 & \textbf{8.27G} & 24.56M \\
    ViT-B/32\cite{dosovitskiy2020image} & 0.8063 & 0.8687 & \multicolumn{1}{c|}{0.6195} & 0.6171 & 0.7652 & \multicolumn{1}{c|}{0.4438} & 0.7117 & 0.8170 & 0.5316 & 8.73G & 87.61M \\
    MUSIQ\cite{ke2021musiq} & 0.8349 & 0.8862 & \multicolumn{1}{c|}{0.6496} & 0.6292 & 0.7839 & \multicolumn{1}{c|}{0.4557} & 0.7321 & 0.8350 & 0.5526 & 124.77G & 78.55M \\
    DB-CNN\cite{zhang2018blind} & 0.8154 & 0.8747 & \multicolumn{1}{c|}{0.6278} & 0.6329 & 0.7823 & \multicolumn{1}{c|}{0.4567} & 0.7242 & 0.8285 & 0.5423 & 33.00G & \textbf{15.31M} \\
    HyperIQA\cite{su2020blindly} & 0.8400 & 0.8958 & \multicolumn{1}{c|}{0.6565} & 0.6276 & 0.8087 & \multicolumn{1}{c|}{0.4543} & 0.7338 & 0.8522 & 0.5554 & 8.67G & 27.38M \\
    TReS\cite{golestaneh2022no} & 0.8367 & 0.8973 & \multicolumn{1}{c|}{0.6531} & 0.6366 & 0.8134 & \multicolumn{1}{c|}{0.4620} & 0.7366 & 0.8553 & 0.5575 & 16.77G & 34.46M \\
    Re-IQA\cite{saha2023re} & 0.8187 & 0.8799 & \multicolumn{1}{c|}{0.6312} & 0.6373 & 0.7880 & \multicolumn{1}{c|}{0.4606} & 0.7280 & 0.8339 & 0.5459 & 33.09G & 40.29M \\
    StairIQA\cite{sun2023blind} & 0.8235 & 0.8864 & \multicolumn{1}{c|}{0.6381} & 0.6348 & 0.8006 & \multicolumn{1}{c|}{0.4600} & 0.7291 & 0.8435 & 0.5490 & 10.22G & 33.01M \\ \hline
    PSCR\cite{yuan2023pscr} & 0.8498 & 0.9059 & \multicolumn{1}{c|}{\textbf{-}} & \textbf{-} & \textbf{-} & \multicolumn{1}{c|}{\textbf{-}} & - & - & - & - & - \\ \hline
    AMFF-Net(ours) & \textbf{0.8565} & \textbf{0.9050} & \multicolumn{1}{c|}{\textbf{0.6759}} & \textbf{0.7513} & \textbf{0.8476} & \multicolumn{1}{c|}{\textbf{0.5663}} & \textbf{0.8039} & \textbf{0.8763} & \textbf{0.6211} & 41.48G & 50.30M \\
    \bottomrule
  \end{tabular}
  \end{threeparttable}
\end{table*}

\begin{table*}[!htpb]
  \centering
  \caption{Quantitative results comparison between different blind IQA methods on AIGCIQA2023. For convenient viewing, we also present the average value of each evaluation metric on three tasks in the last three columns of the table.}
  \label{table.2}
  \begin{threeparttable}
  \renewcommand{\arraystretch}{1.0}
  \begin{tabular}{c|cccccccccccc}
    \toprule
    \multirow{3}{*}{Method} & \multicolumn{12}{c}{AIGCIQA2023\cite{wang2023aigciqa2023}} \\ \cline{2-13}
     & \multicolumn{3}{c|}{Quality} & \multicolumn{3}{c|}{Consistency} & \multicolumn{3}{c|}{Authenticity} & \multicolumn{3}{c}{Avg.} \\ \cline{2-13}
     & SRCC & PLCC & \multicolumn{1}{c|}{KRCC} & SRCC & PLCC & \multicolumn{1}{c|}{KRCC} & SRCC & PLCC & \multicolumn{1}{c|}{KRCC} & SRCC & PLCC & KRCC \\ \hline
    ResNet50\cite{he2016deep} & 0.8208 & 0.8408 & \multicolumn{1}{c|}{0.6083} & 0.6997 & 0.6962 & \multicolumn{1}{c|}{0.5038} & 0.7087 & 0.6964 & \multicolumn{1}{c|}{0.5073} & 0.7431 & 0.7445 & 0.5398 \\
    ViT-B/32\cite{dosovitskiy2020image} & 0.7881 & 0.8140 & \multicolumn{1}{c|}{0.5737} & 0.6404 & 0.6413 & \multicolumn{1}{c|}{0.4557} & 0.6975 & 0.6920 & \multicolumn{1}{c|}{0.4991} & 0.7087 & 0.7158 & 0.5095 \\
    MUSIQ\cite{ke2021musiq} & 0.8423 & 0.8596 & \multicolumn{1}{c|}{0.6327} & 0.7620 & 0.7527 & \multicolumn{1}{c|}{0.5591} & 0.7615 & 0.7509 & \multicolumn{1}{c|}{0.5585} & 0.7886 & 0.7878 & 0.5835 \\
    DB-CNN\cite{zhang2018blind} & 0.8339 & 0.8577 & \multicolumn{1}{c|}{0.6240} & 0.6837 & 0.6787 & \multicolumn{1}{c|}{0.4915} & 0.7485 & 0.7436 & \multicolumn{1}{c|}{0.5449} & 0.7554 & 0.7600 & 0.5535 \\
    HyperIQA\cite{su2020blindly} & \textbf{0.8483} & \textbf{0.8689} & \multicolumn{1}{c|}{\textbf{0.6389}} & 0.7541 & 0.7439 & \multicolumn{1}{c|}{0.5517} & 0.7798 & 0.7718 & \multicolumn{1}{c|}{\textbf{0.5766}} & 0.7940 & \textbf{0.7949} & 0.5890 \\
    TReS\cite{golestaneh2022no} & 0.8436 & 0.8666 & \multicolumn{1}{c|}{0.6357} & 0.7292 & 0.7266 & \multicolumn{1}{c|}{0.5331} & 0.7661 & 0.7602 & \multicolumn{1}{c|}{0.5655} & 0.7796 & 0.7845 & 0.5781 \\
    Re-IQA\cite{saha2023re} & 0.8144 & 0.8317 & \multicolumn{1}{c|}{0.5980} & 0.6430 & 0.6355 & \multicolumn{1}{c|}{0.4564} & 0.7224 & 0.7110 & \multicolumn{1}{c|}{0.5214} & 0.7266 & 0.7261 & 0.5253 \\
    StairIQA\cite{sun2023blind} & 0.8186 & 0.8450 & \multicolumn{1}{c|}{0.6063} & 0.6641 & 0.6625 & \multicolumn{1}{c|}{0.4748} & 0.7155 & 0.7131 & \multicolumn{1}{c|}{0.5151} & 0.7328 & 0.7402 & 0.5321 \\ \hline
    PSCR\cite{yuan2023pscr} & 0.8371 & 0.8588 & \multicolumn{1}{c|}{\textbf{-}} & 0.7465 & 0.7379 & \multicolumn{1}{c|}{\textbf{-}} & \textbf{0.7828} & \textbf{0.7750} & \multicolumn{1}{c|}{-} & 0.7888 & 0.7906 & - \\ \hline
    AMFF-Net(ours) & 0.8409 & 0.8537 & \multicolumn{1}{c|}{0.6310} & \textbf{0.7782} & \textbf{0.7638} & \multicolumn{1}{c|}{\textbf{0.5747}} & 0.7749 & 0.7643 & \multicolumn{1}{c|}{0.5684} & \textbf{0.7980} & 0.7939 & \textbf{0.5914} \\
    \bottomrule
    \end{tabular}
  \end{threeparttable}
\end{table*}

\begin{table*}[!htpb]
  \centering
  \caption{Quantitative results comparison between different blind IQA methods on PKU-I2IQA. For convenient viewing, we also present the average value of each evaluation metric on three tasks in the last three columns of the table.}
  \label{table.3}
  \begin{threeparttable}
  \renewcommand{\arraystretch}{1.0}
  \begin{tabular}{c|cccccccccccc}
    \toprule
    \multirow{3}{*}{Method} & \multicolumn{12}{c}{PKUI2IQA\cite{yuan2023pku}} \\ \cline{2-13}
     & \multicolumn{3}{c|}{Quality} & \multicolumn{3}{c|}{Consistency} & \multicolumn{3}{c|}{Authenticity} & \multicolumn{3}{c}{Avg.} \\ \cline{2-13}
     & SRCC & PLCC & \multicolumn{1}{c|}{KRCC} & SRCC & PLCC & \multicolumn{1}{c|}{KRCC} & SRCC & PLCC & \multicolumn{1}{c|}{KRCC} & SRCC & PLCC & KRCC \\ \hline
    ResNet50\cite{he2016deep} & 0.6219 & 0.6328 & \multicolumn{1}{c|}{0.4440} & 0.6601 & 0.6511 & \multicolumn{1}{c|}{0.4792} & 0.4932 & 0.5211 & \multicolumn{1}{c|}{0.3437} & 0.5917 & 0.6017 & 0.4223 \\
    ViT-B/32\cite{dosovitskiy2020image} & 0.5151 & 0.5197 & \multicolumn{1}{c|}{0.3598} & 0.5507 & 0.5389 & \multicolumn{1}{c|}{0.3862} & 0.4226 & 0.4485 & \multicolumn{1}{c|}{0.2937} & 0.4962 & 0.5024 & 0.3465 \\
    MUSIQ\cite{ke2021musiq} & 0.6408 & 0.6410 & \multicolumn{1}{c|}{0.4596} & 0.6379 & 0.6531 & \multicolumn{1}{c|}{0.4599} & 0.6354 & 0.6505 & \multicolumn{1}{c|}{0.4571} & 0.6380 & 0.6482 & 0.4589 \\
    DB-CNN\cite{zhang2018blind} & 0.5975 & 0.5970 & \multicolumn{1}{c|}{0.4247} & 0.6083 & 0.5925 & \multicolumn{1}{c|}{0.4345} & 0.5667 & 0.5820 & \multicolumn{1}{c|}{0.4024} & 0.5908 & 0.5905 & 0.4205 \\
    HyperIQA\cite{su2020blindly} & 0.6849 & 0.6955 & \multicolumn{1}{c|}{0.4988} & 0.7239 & 0.7062 & \multicolumn{1}{c|}{0.5378} & 0.6596 & 0.6902 & \multicolumn{1}{c|}{0.4794} & 0.6894 & 0.6973 & 0.5053 \\
    TReS\cite{golestaneh2022no} & 0.6374 & 0.6427 & \multicolumn{1}{c|}{0.4572} & 0.6480 & 0.6456 & \multicolumn{1}{c|}{0.4690} & 0.6003 & 0.6393 & \multicolumn{1}{c|}{0.4311} & 0.6286 & 0.6425 & 0.4524 \\
    Re-IQA\cite{saha2023re} & 0.5996 & 0.6106 & \multicolumn{1}{c|}{0.4248} & 0.5705 & 0.5690 & \multicolumn{1}{c|}{0.4029} & 0.5509 & 0.5787 & \multicolumn{1}{c|}{0.3884} & 0.5737 & 0.5861 & 0.4054 \\
    StairIQA\cite{sun2023blind} & 0.5855 & 0.6038 & \multicolumn{1}{c|}{0.4151} & 0.5739 & 0.5720 & \multicolumn{1}{c|}{0.4048} & 0.5535 & 0.5879 & \multicolumn{1}{c|}{0.3927} & 0.5709 & 0.5879 & 0.4042 \\ \hline
    AMFF-Net(ours) & \textbf{0.7065} & \textbf{0.7174} & \multicolumn{1}{c|}{\textbf{0.5169}} & \textbf{0.7796} & \textbf{0.7708} & \multicolumn{1}{c|}{\textbf{0.5921}} & \textbf{0.6836} & \textbf{0.7206} & \multicolumn{1}{c|}{\textbf{0.5002}} & \textbf{0.7232} & \textbf{0.7363} & \textbf{0.5364} \\
    \bottomrule
    \end{tabular}
  \end{threeparttable}
\end{table*}

\subsection{Performance Comparison}
\label{ssec:subhead}
\subsubsection{Prediction Ability Comparison}
In this study, we compare our proposed AMFF-Net with nine state-of-the-art blind IQA methods, including ResNet50 \cite{he2016deep}, ViT-B/32 \cite{dosovitskiy2020image}, MUSIQ\cite{ke2021musiq}, DB-CNN \cite{zhang2018blind}, HyperIQA \cite{su2020blindly}, TReS\cite{golestaneh2022no}, Re-IQA \cite{saha2023re}, StairIQA \cite{sun2023blind}, and PSCR \cite{yuan2023pscr}. Among these methods, the first eight methods are designed for NSIs, while the last one is a method specifically designed for AGIs. All these methods, except PSCR, are re-trained on the three AGI quality assessment databases using their default settings. Since PSCR does not release the source code, we directly extract results from its original paper, in which only partial results on AGIQA-3K and AIGCIQA2023 are reported. Table \ref{table.1}, Table \ref{table.2}, and Table \ref{table.3} tabulate the quantitative results of our proposed AMFF-Net and other methods on three databases. The results are the median values of 10 trials, in which a random train-test split is conducted according to the settings in Section \ref{sssec:4.1}. For the convenience of comparison, the best results are marked in bold.

\begin{figure*}[!htpb]%
  \centering
  \includegraphics[width=2.0\columnwidth]{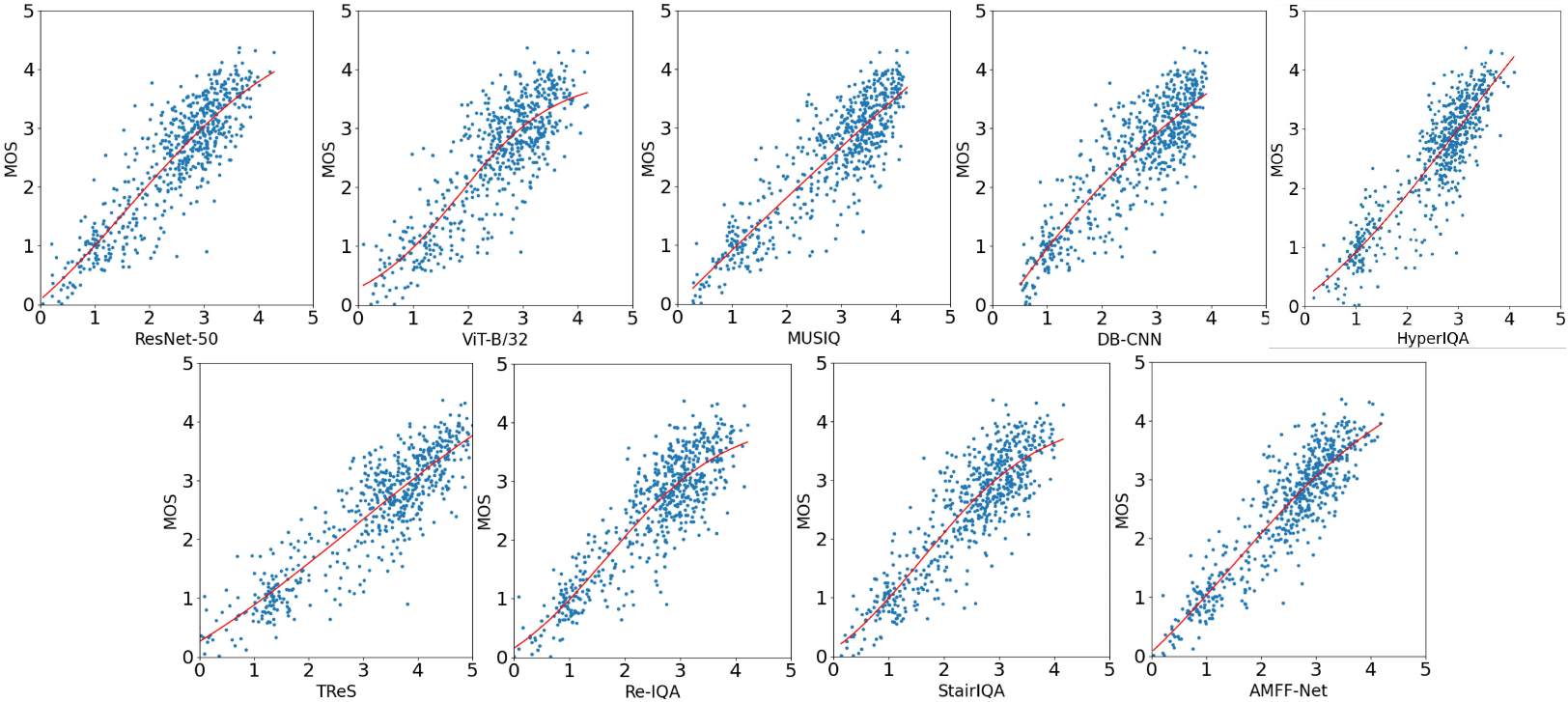}\\
  \caption{Scatter plots of different IQA methods tested on the AGIQA-3K database. Due to the space limitation, we only show the predictions of visual quality.}\label{Fig.5}
\end{figure*}

From the tables, we have the following observations. First, AMFF-Net outperforms these competing methods on AGIQA-3K and PKU-I2IQA databases. For example, AMFF-Net is ahead of the second best method (i.e., TReS) by approximately 2.367\% and 18.018\% in terms of SRCC when evaluating the visual quality and consistency on AGIQA-3K, respectively. It also achieves a performance gain by approximately 3.154\%, 7.694\%, and 3.639\% than the second best method (i.e., HyperIQA) in terms of SRCC when evaluating visual quality, consistency, and authenticity on PKU-I2IQA. Second, our AMFF-Net ranks the second when evaluating visual quality and authenticity on AIGCIQA2023 and is slightly inferior to the best method (i.e., HyperIQA) by approximately 0.872\% and 0.628\%. A possible reason for this is that, the images (e.g., generated by the Lafite model \cite{zhou2021lafite}) in the AIGCIQA2023 possess too similar characteristics, while our method has limited capability to discriminate fine-grained differences between images. Despite this, it still exhibits superior performance than other methods in evaluating consistency and achieves the best average result of three dimensions, with 3.196\% SRCC advantages over HyperIQA. Third, among all selected NSI oriented methods, HyperIQA performance best. A possible reason for this is that, HyperIQA evaluates the image quality in a content-aware manner using a hyper network, contributing to understand the semantic distortions, which get more attention during subjective rating of AGIs. Fourth, the specifically designed method (PSCR) for AGIs generally performs better than most traditional NSI-oriented IQA methods. This may be attributed to the fact that it compares two AGIs generated by the same text prompts during the network design. Nevertheless, it is inferior to our proposed AMFF-Net in most cases.
To compare these methods more intuitively, Fig. \ref{Fig.5} presents the scatter plots of different IQA methods on AGIQA-3K database. As seen, our AMFF-Net produces more consistent predictions with subjective ratings than competing methods.

We also compare our AMFF-Net with competing methods in terms of the floating-point operations (FLOPs) and the number of parameters (\#Params). As shown in the last two columns of Table \ref{table.1}, AMFF-Net has 41.48G of FLOPs and 50.30M of \#Params, ranking eighth and seventh among ten competing methods, respectively. This indicates that, compared to its competitors, our AMFF-Net does not exhibit competitive advantage in these two aspects. One possible reason for this could be that AMFF-Net requires multi-scale images as input and needs to process features from three different scales in the inference stage. Despite this, our proposed AMFF-Net is more competent for the quality assessment tasks in terms of prediction accuracy.

\begin{table*}[!htpb]
  \centering
  \caption{Cross-validation results of different blind IQA methods.}
  \label{table.4}
  \renewcommand{\arraystretch}{1.0}
  \begin{tabular}{c|cccccc|cccccc}
    \toprule
    \multirow{3}{*}{Method} & \multicolumn{6}{c|}{AIGCIQA2023(Train) $\rightarrow$ AGIQA-3K(Test)} & \multicolumn{6}{c}{AGIQA-3K(Train) $\rightarrow$ AIGCIQA2023(Test)} \\ \cline{2-13}
     & \multicolumn{3}{c|}{Quality} & \multicolumn{3}{c|}{Consistency} & \multicolumn{3}{c|}{Quality} & \multicolumn{3}{c}{Consistency} \\ \cline{2-13}
     & SRCC & PLCC & \multicolumn{1}{c|}{KRCC} & SRCC & PLCC & KRCC & SRCC & PLCC & \multicolumn{1}{c|}{KRCC} & SRCC & PLCC & KRCC \\ \hline
     RestNet50\cite{he2016deep} & 0.576 & 0.612 & \multicolumn{1}{c|}{0.397} & 0.473 & 0.523 & 0.326 & 0.599 & 0.609 & \multicolumn{1}{c|}{0.413} & 0.432 & 0.435 & 0.305 \\
     ViT-B/32\cite{dosovitskiy2020image} & 0.509 & 0.571 & \multicolumn{1}{c|}{0.350} & 0.434 & 0.496 & 0.302 & 0.517 & 0.529 & \multicolumn{1}{c|}{0.357} & 0.381 & 0.390 & 0.259 \\
     MUSIQ\cite{ke2021musiq} & 0.635 & 0.673 & \multicolumn{1}{c|}{0.445} & 0.394 & 0.437 & 0.265 & 0.650 & 0.643 & \multicolumn{1}{c|}{0.444} & 0.525 & 0.515 & 0.359 \\
     DB-CNN\cite{zhang2018blind} & 0.627 & 0.688 & \multicolumn{1}{c|}{0.442} & 0.390 & 0.435 & 0.263 & 0.654 & 0.664 & \multicolumn{1}{c|}{0.455} & 0.470 & 0.460 & 0.317 \\
     HyperIQA\cite{su2020blindly} & \textbf{0.657} & 0.692 & \multicolumn{1}{c|}{0.443} & 0.418 & 0.465 & 0.280 & 0.669 & \textbf{0.672} & \multicolumn{1}{c|}{0.472} & 0.464 & 0.431 & 0.314 \\
     TReS\cite{golestaneh2022no} & 0.646 & \textbf{0.702} & \multicolumn{1}{c|}{0.451} & 0.445 & 0.488 & 0.303 & 0.650 & 0.654 & \multicolumn{1}{c|}{0.440} & 0.505 & 0.483 & 0.346 \\
     Re-IQA\cite{saha2023re} & 0.473 & 0.352 & \multicolumn{1}{c|}{0.325} & 0.243 & 0.154 & 0.165 & 0.654 & 0.654 & \multicolumn{1}{c|}{0.441} & 0.479 & 0.484 & 0.323 \\ \hline
     AMFF-Net(ours) & 0.654 & 0.695 & \multicolumn{1}{c|}{\textbf{0.459}} & \textbf{0.554} & \textbf{0.624} & \textbf{0.385} & \textbf{0.678} & 0.669 & \multicolumn{1}{c|}{\textbf{0.474}} & \textbf{0.546} & \textbf{0.549} & \textbf{0.381} \\
    \bottomrule
    \end{tabular}
\end{table*}

\begin{table*}[!htpb]
  \centering
  \caption{Results of ablation experiments on three databases. Due to the space limitation, only the SRCC values are presented here.}
  \label{table.5}
  \renewcommand{\arraystretch}{1.0}
  \begin{tabular}{c|cc|ccc|ccc}
    \toprule
    \multirow{2}{*}{Method} & \multicolumn{2}{c|}{AGIQA-3K\cite{li2023agiqa}} & \multicolumn{3}{c|}{AIGCIQA2023\cite{wang2023aigciqa2023}} & \multicolumn{3}{c}{PKU-I2IQA\cite{yuan2023pku}} \\ \cline{2-9}
     & Quality & Consistency & Quality & Consistency & Authenticity & Quality & Consistency & Authenticity \\ \hline
    w/o MSI & 0.855 & 0.748 & 0.827 & 0.774 & 0.766 & 0.670 & \textbf{0.781} & 0.648 \\
    w/o AFF & 0.852 & 0.739 & 0.835 & 0.768 & 0.774 & 0.694 & \textbf{0.781} & 0.680 \\
    AMFF-Net & \textbf{0.856} & \textbf{0.751} & \textbf{0.841} & \textbf{0.778} & \textbf{0.775} & \textbf{0.706} & 0.780 & \textbf{0.684} \\
    \bottomrule
    \end{tabular}
\end{table*}

\subsubsection{Generalization Ability Comparison}

Apart from prediction ability, generalization ability is another crucial factor for a blind IQA method. In this section, we further compare these selected methods in terms of generalization ability by conducting cross-validation experiments. Since the images in AGIQA-3K and AIGCIQA2023 are generated by Text-to-Image generative models, while those in PKU-I2IQA are generated by Image-to-Image generative models, we select AGIQA-3K and AIGCIQA2023 for the experiments. Specifically, each method is trained on one database and tested on the other database when evaluating a specific quality of AGIs, i.e., visual quality and content consistency. Table \ref{table.4} presents the experimental results in the form of three evaluation metrics. Here, we do not consider StairIQA as it utilizes mixed databases for training, which is unfair for the cross-validation comparison. From the table, we can observe that our proposed AMFF-Net outperforms the competing blind IQA methods by a large margin, with higher SRCC, PLCC, and SRCC values in most cases. This indicates its higher generalization ability.

\subsection{Ablation Experiments}
\label{sssec:subsubhead}
\subsubsection{Effectiveness of Each Component}
In this study, the proposed AMFF-Net utilizes the multi-scale input strategy (MSI) and adaptive feature fusion (AFF) block for accurate quality prediction. Here, we conduct some ablation studies to investigate the effectiveness of the MSI strategy and the AFF block. The experimental settings are the same as the main experiment, and the median results of 10 trials are reported. Due to the space limitation, only the SRCC scores are given, as shown in Table \ref{table.5}. The ``w/o MSI'' denotes the learned IQA model when using three original-sized images as the inputs. The ``w/o AFF'' is the learned IQA model when directly adding the multi-scale representations, instead of using the AFF block for adaptive fusion. From the table, it is clear that the absence of either the MSI strategy or the AFF block will degrade the prediction performance. For example, the removal of the MSI strategy and the AFF block leads to a SRCC drop of 0.713\% and 1.665\% respectively when evaluating visual quality on AIGCIQA2023. Similarly, AMFF-Net has a SRCC drop of 0.585\% and 5.263\% if we do not use the MSI strategy and the AFF block separately when evaluating the authenticity on PKU-I2IQA. These results show that both the MSI strategy and the AFF block play a positive role in evaluating AGIs, contributing to achieving accurate predictions for AMFF-Net.

\subsubsection{Effectiveness of the Similarity Metric}
\begin{figure}[!htpb]
  \centering
  \includegraphics[width=\columnwidth]{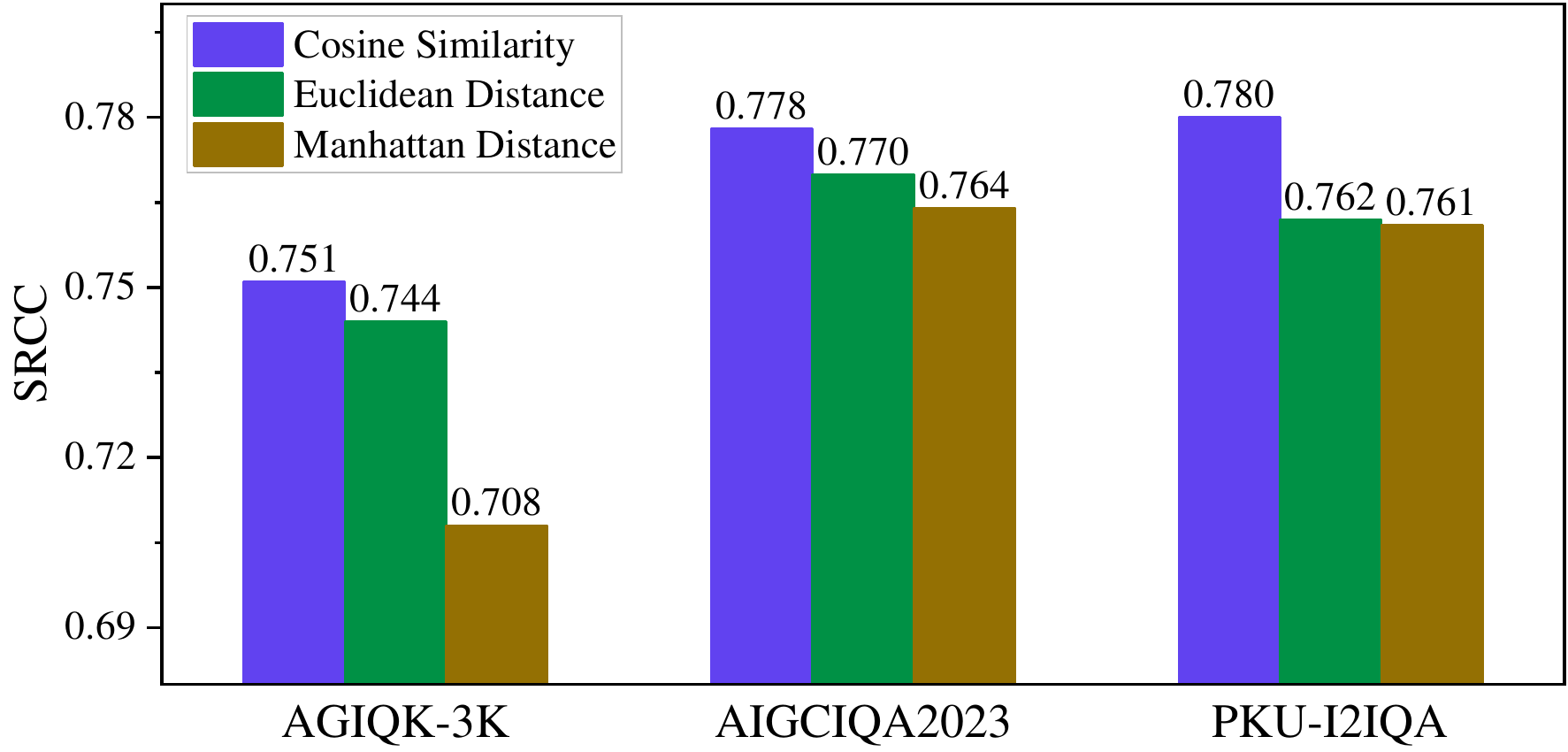}\\
  \caption{SRCC results of three similarity metrics for consistency prediction on three databases.}\label{Fig.6}
\end{figure}

In Eq. \eqref{Eq.3}, we choose cosine function to measure the similarity between text and image features. Here, we investigate its effectiveness on the AGI quality task by comparing it with two other similarity metrics: Euclidean distance and Manhattan distance. Fig. \ref{Fig.6} shows the the SRCC results on three databases. It can be observed that cosine function achieves better results than Euclidean distance and Manhattan distance.

\subsection{Limitations and Future Works}
\label{sec:typestyle}
Although our proposed AMFF-Net has demonstrated superiority over competing methods in quality assessment of AGIs, it still faces certain limitations. On the one hand, the prediction accuracy of AMFF-Net can be further improved. On the other hand, the computational complexity of AMFF-Net needs to be reduced. To update the proposed AMFF-Net, future works can be carried out from the following directions. Firstly, to improve the prediction accuracy, we will further strengthen the interaction between text features and image features, so that the model can get more discriminative features. Secondly, to reduce the computational complexity and accelerate inference speed, we will optimize the model structure by lightweight modules and design parallel computation schemes to simultaneously process different scaled inputs.

\section{Conclusion}
\label{sec:typestyle}

AGI quality assessment has recently emerged as a new and important topic, requiring comprehensive evaluation from multiple dimensions. In this paper, we propose a simple yet effective blind IQA network, termed AMFF-Net, for AGIs. Different from existing methods that only evaluate the images from the perspective of ``visual quality'', AMFF-Net utilizes a multi-task framework and aims to evaluate AGI quality from three dimensions, i.e., ``visual quality'', ``authenticity'', and ``consistency''. Specifically, considering that ``visual quality'' and ``authenticity'' are characterized by both local and global aspects, AMFF-Net utilizes a multi-scale input (MSI) strategy to capture image details at different levels of granularity. After that, an adaptive feature fusion (AFF) block is used to adaptively multi-scale features. To evaluate the content consistency, the similarity between semantic features of text prompts and AGI is computed. Through the cooperation of the MSI strategy and the AFF block, our AMFF-Net performs better than nine state-of-the-art blind IQA methods on three publicly available AGI quality assessment databases. In addition, ablations experiments demonstrate the effectiveness of the underlying concepts of the MSI strategy and the AFF block.

\bibliographystyle{IEEEtran}
\bibliography{refs}

\begin{thebibliography}{10}
\providecommand{\url}[1]{#1}
\csname url@samestyle\endcsname
\providecommand{\newblock}{\relax}
\providecommand{\bibinfo}[2]{#2}
\providecommand{\BIBentrySTDinterwordspacing}{\spaceskip=0pt\relax}
\providecommand{\BIBentryALTinterwordstretchfactor}{4}
\providecommand{\BIBentryALTinterwordspacing}{\spaceskip=\fontdimen2\font plus
\BIBentryALTinterwordstretchfactor\fontdimen3\font minus \fontdimen4\font\relax}
\providecommand{\BIBforeignlanguage}[2]{{%
\expandafter\ifx\csname l@#1\endcsname\relax
\typeout{** WARNING: IEEEtran.bst: No hyphenation pattern has been}%
\typeout{** loaded for the language `#1'. Using the pattern for}%
\typeout{** the default language instead.}%
\else
\language=\csname l@#1\endcsname
\fi
#2}}
\providecommand{\BIBdecl}{\relax}
\BIBdecl

\bibitem{gan2023web}
W.~Gan, Z.~Ye, S.~Wan, and P.~S. Yu, ``Web 3.0: The future of internet,'' \emph{arXiv preprint arXiv:2304.06032}, 2023.

\bibitem{frolov2021adversarial}
S.~Frolov, T.~Hinz, F.~Raue, J.~Hees, and A.~Dengel, ``Adversarial text-to-image synthesis: A review,'' \emph{Neural Networks}, vol. 144, pp. 187--209, 2021.

\bibitem{li2023agiqa}
C.~Li, Z.~Zhang, H.~Wu, W.~Sun, X.~Min, X.~Liu, G.~Zhai, and W.~Lin, ``Agiqa-3k: An open database for ai-generated image quality assessment,'' \emph{IEEE Transactions on Circuits and Systems for Video Technology}, accepted, in press, DOI: 10.1109/TCSVT.2023.3319020, 2023.

\bibitem{goring2023appeal}
S.~G{\"o}ring, R.~R.~R. Rao, R.~Merten, and A.~Raake, ``Appeal and quality assessment for ai-generated images,'' in \emph{2023 15th International Conference on Quality of Multimedia Experience (QoMEX)}.\hskip 1em plus 0.5em minus 0.4em\relax IEEE, 2023, pp. 115--118.

\bibitem{lan2023multilevel}
X.~Lan, M.~Zhou, X.~Xu, X.~Wei, X.~Liao, H.~Pu, J.~Luo, T.~Xiang, B.~Fang, and Z.~Shang, ``Multilevel feature fusion for end-to-end blind image quality assessment,'' \emph{IEEE Transactions on Broadcasting}, vol.~69, no.~3, pp. 801--811, 2023.

\bibitem{yue2024subjective}
G.~Yue, H.~Wu, W.~Yan, T.~Zhou, H.~Liu, and W.~Zhou, ``Subjective and objective quality assessment of multi-attribute retouched face images,'' \emph{IEEE Transactions on Broadcasting}, accepted, in press, DOI: 10.1109/TBC.2024.3374043, 2024.

\bibitem{1037556601}
B.~Hu, T.~Zhao, J.~Zheng, Y.~Zhang, L.~Li, W.~Li, and X.~Gao, ``Blind image quality assessment with coarse-grained perception construction and fine-grained interaction learning,'' \emph{IEEE Transactions on Broadcasting}, accepted, in press, DOI:10.1109/TBC.2023.3342696, 2023.

\bibitem{yue2022improving}
G.~Yue, D.~Cheng, H.~Wu, Q.~Jiang, and T.~Wang, ``Improving iqa performance based on deep mutual learning,'' in \emph{2022 IEEE International Conference on Image Processing (ICIP)}.\hskip 1em plus 0.5em minus 0.4em\relax IEEE, 2022, pp. 2182--2186.

\bibitem{lang2023full}
S.~Lang, X.~Liu, M.~Zhou, J.~Luo, H.~Pu, X.~Zhuang, J.~Wang, X.~Wei, T.~Zhang, Y.~Feng \emph{et~al.}, ``A full-reference image quality assessment method via deep meta-learning and conformer,'' \emph{IEEE Transactions on Broadcasting}, accepted, in press, DOI: 10.1109/TBC.2023.3308349, 2023.

\bibitem{10337739}
G.~Yue, H.~Wu, Q.~Jiang, T.~Zhou, W.~Yan, and T.~Wang, ``Perceptual quality assessment of retouched face images,'' \emph{IEEE Transactions on Multimedia}, accepted, in press, DOI: 10.1109/TMM.2023.3338412, 2023.

\bibitem{kim2017deep}
J.~Kim, H.~Zeng, D.~Ghadiyaram, S.~Lee, L.~Zhang, and A.~C. Bovik, ``Deep convolutional neural models for picture-quality prediction: Challenges and solutions to data-driven image quality assessment,'' \emph{IEEE Signal Processing Magazine}, vol.~34, no.~6, pp. 130--141, 2017.

\bibitem{hu2023blind}
B.~Hu, G.~Zhu, L.~Li, J.~Gan, W.~Li, and X.~Gao, ``Blind image quality index with cross-domain interaction and cross-scale integration,'' \emph{IEEE Transactions on Multimedia}, accepted, in press, DOI: 10.1109/TMM.2023.3303725, 2023.

\bibitem{bosse2017deep}
S.~Bosse, D.~Maniry, K.-R. M{\"u}ller, T.~Wiegand, and W.~Samek, ``Deep neural networks for no-reference and full-reference image quality assessment,'' \emph{IEEE Transactions on Image Processing}, vol.~27, no.~1, pp. 206--219, 2017.

\bibitem{10440553}
L.~Zheng, Y.~Luo, Z.~Zhou, J.~Ling, and G.~Yue, ``Cdinet: Content distortion interaction network for blind image quality assessment,'' \emph{IEEE Transactions on Multimedia}, accepted, in press, DOI: 10.1109/TMM.2024.3360697, 2024.

\bibitem{ma2017end}
K.~Ma, W.~Liu, K.~Zhang, Z.~Duanmu, Z.~Wang, and W.~Zuo, ``End-to-end blind image quality assessment using deep neural networks,'' \emph{IEEE Transactions on Image Processing}, vol.~27, no.~3, pp. 1202--1213, 2017.

\bibitem{10464346}
T.~Zhou, S.~Tan, B.~Zhao, and G.~Yue, ``Multitask deep neural network with knowledge-guided attention for blind image quality assessment,'' \emph{IEEE Transactions on Circuits and Systems for Video Technology}, accepted, in press, DOI: 10.1109/TCSVT.2024.3375344, 2024.

\bibitem{song2022knowledge}
T.~Song, L.~Li, J.~Wu, Y.~Yang, Y.~Li, Y.~Guo, and G.~Shi, ``Knowledge-guided blind image quality assessment with few training samples,'' \emph{IEEE Transactions on Multimedia}, vol.~25, pp. 8145--8156, 2023.

\bibitem{golestaneh2022no}
S.~A. Golestaneh, S.~Dadsetan, and K.~M. Kitani, ``No-reference image quality assessment via transformers, relative ranking, and self-consistency,'' in \emph{Proceedings of the IEEE/CVF Winter Conference on Applications of Computer Vision}, 2022, pp. 1220--1230.

\bibitem{pan2022vcrnet}
Z.~Pan, F.~Yuan, J.~Lei, Y.~Fang, X.~Shao, and S.~Kwong, ``Vcrnet: Visual compensation restoration network for no-reference image quality assessment,'' \emph{IEEE Transactions on Image Processing}, vol.~31, pp. 1613--1627, 2022.

\bibitem{su2020blindly}
S.~Su, Q.~Yan, Y.~Zhu, C.~Zhang, X.~Ge, J.~Sun, and Y.~Zhang, ``Blindly assess image quality in the wild guided by a self-adaptive hyper network,'' in \emph{Proceedings of the IEEE/CVF Conference on Computer Vision and Pattern Recognition}, 2020, pp. 3667--3676.

\bibitem{yang2019sgdnet}
S.~Yang, Q.~Jiang, W.~Lin, and Y.~Wang, ``Sgdnet: An end-to-end saliency-guided deep neural network for no-reference image quality assessment,'' in \emph{Proceedings of the 27th ACM International Conference on Multimedia}, 2019, pp. 1383--1391.

\bibitem{you2022attention}
J.~You and J.~Korhonen, ``Attention integrated hierarchical networks for no-reference image quality assessment,'' \emph{Journal of Visual Communication and Image Representation}, vol.~82, p. 103399, 2022.

\bibitem{ou2021novel}
F.-Z. Ou, Y.-G. Wang, J.~Li, G.~Zhu, and S.~Kwong, ``A novel rank learning based no-reference image quality assessment method,'' \emph{IEEE Transactions on Multimedia}, vol.~24, pp. 4197--4211, 2022.

\bibitem{yue2022semi}
G.~Yue, D.~Cheng, L.~Li, T.~Zhou, H.~Liu, and T.~Wang, ``Semi-supervised authentically distorted image quality assessment with consistency-preserving dual-branch convolutional neural network,'' \emph{IEEE Transactions on Multimedia}, vol.~25, pp. 6499--6511, 2023.

\bibitem{yang2023multi}
Z.~Yang, L.~Li, Y.~Yang, Y.~Li, and W.~Lin, ``Multi-level transitional contrast learning for personalized image aesthetics assessment,'' \emph{IEEE Transactions on Multimedia}, vol.~26, pp. 1944--1956, 2024.

\bibitem{lin2019kadid}
H.~Lin, V.~Hosu, and D.~Saupe, ``Kadid-10k: A large-scale artificially distorted iqa database,'' in \emph{2019 Eleventh International Conference on Quality of Multimedia Experience (QoMEX)}.\hskip 1em plus 0.5em minus 0.4em\relax IEEE, 2019, pp. 1--3.

\bibitem{lin2018koniq}
------, ``Koniq-10k: Towards an ecologically valid and large-scale iqa database,'' \emph{arXiv preprint arXiv:1803.08489}, 2018.

\bibitem{yuan2023pku}
J.~Yuan, X.~Cao, C.~Li, F.~Yang, J.~Lin, and X.~Cao, ``Pku-i2iqa: An image-to-image quality assessment database for ai generated images,'' \emph{arXiv preprint arXiv:2311.15556}, 2023.

\bibitem{heusel2017gans}
M.~Heusel, H.~Ramsauer, T.~Unterthiner, B.~Nessler, and S.~Hochreiter, ``Gans trained by a two time-scale update rule converge to a local nash equilibrium,'' \emph{Advances in Neural Information Processing Systems}, vol.~30, 2017.

\bibitem{binkowski2018demystifying}
M.~Bi{\'n}kowski, D.~J. Sutherland, M.~Arbel, and A.~Gretton, ``Demystifying mmd gans,'' \emph{arXiv preprint arXiv:1801.01401}, 2018.

\bibitem{wang2023aigciqa2023}
J.~Wang, H.~Duan, J.~Liu, S.~Chen, X.~Min, and G.~Zhai, ``Aigciqa2023: A large-scale image quality assessment database for ai generated images: from the perspectives of quality, authenticity and correspondence,'' \emph{arXiv preprint arXiv:2307.00211}, 2023.

\bibitem{yuan2024tier}
J.~Yuan, X.~Cao, J.~Che, Q.~Wang, S.~Liang, W.~Ren, J.~Lin, and X.~Cao, ``Tier: Text and image encoder-based regression for aigc image quality assessment,'' \emph{arXiv preprint arXiv:2401.03854}, 2024.

\bibitem{radford2021learning}
A.~Radford, J.~W. Kim, C.~Hallacy, A.~Ramesh, G.~Goh, S.~Agarwal, G.~Sastry, A.~Askell, P.~Mishkin, J.~Clark \emph{et~al.}, ``Learning transferable visual models from natural language supervision,'' in \emph{International Conference on Machine Learning}.\hskip 1em plus 0.5em minus 0.4em\relax PMLR, 2021, pp. 8748--8763.

\bibitem{zhou2023perception}
M.~Zhou, L.~Chen, X.~Wei, X.~Liao, Q.~Mao, H.~Wang, H.~Pu, J.~Luo, T.~Xiang, and B.~Fang, ``Perception-oriented u-shaped transformer network for 360-degree no-reference image quality assessment,'' \emph{IEEE Transactions on Broadcasting}, vol.~69, no.~2, pp. 396--405, 2023.

\bibitem{yue2019blind}
G.~Yue, C.~Hou, W.~Yan, L.~K. Choi, T.~Zhou, and Y.~Hou, ``Blind quality assessment for screen content images via convolutional neural network,'' \emph{Digital Signal Processing}, vol.~91, pp. 21--30, 2019.

\bibitem{he2016deep}
K.~He, X.~Zhang, S.~Ren, and J.~Sun, ``Deep residual learning for image recognition,'' in \emph{Proceedings of the IEEE Conference on Computer Vision and Pattern Recognition}, 2016, pp. 770--778.

\bibitem{ke2021musiq}
J.~Ke, Q.~Wang, Y.~Wang, P.~Milanfar, and F.~Yang, ``Musiq: Multi-scale image quality transformer,'' in \emph{Proceedings of the IEEE/CVF International Conference on Computer Vision}, 2021, pp. 5148--5157.

\bibitem{wu2014no}
Q.~Wu, H.~Li, K.~N. Ngan, B.~Zeng, and M.~Gabbouj, ``No reference image quality metric via distortion identification and multi-channel label transfer,'' in \emph{2014 IEEE International Symposium on Circuits and Systems (ISCAS)}.\hskip 1em plus 0.5em minus 0.4em\relax IEEE, 2014, pp. 530--533.

\bibitem{yan2019naturalness}
B.~Yan, B.~Bare, and W.~Tan, ``Naturalness-aware deep no-reference image quality assessment,'' \emph{IEEE Transactions on Multimedia}, vol.~21, no.~10, pp. 2603--2615, 2019.

\bibitem{sun2023blind}
W.~Sun, X.~Min, D.~Tu, S.~Ma, and G.~Zhai, ``Blind quality assessment for in-the-wild images via hierarchical feature fusion and iterative mixed database training,'' \emph{IEEE Journal of Selected Topics in Signal Processing}, vol.~17, no.~6, pp. 1178--1192, 2023.

\bibitem{salimans2016improved}
T.~Salimans, I.~Goodfellow, W.~Zaremba, V.~Cheung, A.~Radford, and X.~Chen, ``Improved techniques for training gans,'' \emph{Advances in Neural Information Processing Systems}, vol.~29, 2016.

\bibitem{zhang2023perceptual}
Z.~Zhang, C.~Li, W.~Sun, X.~Liu, X.~Min, and G.~Zhai, ``A perceptual quality assessment exploration for aigc images,'' \emph{arXiv preprint arXiv:2303.12618}, 2023.

\bibitem{yuan2023pscr}
J.~Yuan, X.~Cao, L.~Cao, J.~Lin, and X.~Cao, ``Pscr: Patches sampling-based contrastive regression for aigc image quality assessment,'' \emph{arXiv preprint arXiv:2312.05897}, 2023.

\bibitem{kirstain2024pick}
Y.~Kirstain, A.~Polyak, U.~Singer, S.~Matiana, J.~Penna, and O.~Levy, ``Pick-a-pic: An open dataset of user preferences for text-to-image generation,'' \emph{Advances in Neural Information Processing Systems}, vol.~36, 2024.

\bibitem{lu2024llmscore}
Y.~Lu, X.~Yang, X.~Li, X.~E. Wang, and W.~Y. Wang, ``Llmscore: Unveiling the power of large language models in text-to-image synthesis evaluation,'' \emph{Advances in Neural Information Processing Systems}, vol.~36, 2024.

\bibitem{wu2023human}
X.~Wu, K.~Sun, F.~Zhu, R.~Zhao, and H.~Li, ``Human preference score: Better aligning text-to-image models with human preference,'' in \emph{Proceedings of the IEEE/CVF International Conference on Computer Vision}, 2023, pp. 2096--2105.

\bibitem{abdelfattah2023cdul}
R.~Abdelfattah, Q.~Guo, X.~Li, X.~Wang, and S.~Wang, ``Cdul: Clip-driven unsupervised learning for multi-label image classification,'' in \emph{Proceedings of the IEEE/CVF International Conference on Computer Vision}, 2023, pp. 1348--1357.

\bibitem{teng2021global}
Z.~Teng, Y.~Duan, Y.~Liu, B.~Zhang, and J.~Fan, ``Global to local: Clip-lstm-based object detection from remote sensing images,'' \emph{IEEE Transactions on Geoscience and Remote Sensing}, vol.~60, pp. 1--13, 2021.

\bibitem{sain2023clip}
A.~Sain, A.~K. Bhunia, P.~N. Chowdhury, S.~Koley, T.~Xiang, and Y.-Z. Song, ``Clip for all things zero-shot sketch-based image retrieval, fine-grained or not,'' in \emph{Proceedings of the IEEE/CVF Conference on Computer Vision and Pattern Recognition}, 2023, pp. 2765--2775.

\bibitem{wang2023exploring}
J.~Wang, K.~C. Chan, and C.~C. Loy, ``Exploring clip for assessing the look and feel of images,'' in \emph{Proceedings of the AAAI Conference on Artificial Intelligence}, vol.~37, no.~2, 2023, pp. 2555--2563.

\bibitem{pan2023quality}
W.~Pan, Z.~Yang, D.~Liu, C.~Fang, Y.~Zhang, and P.~Dai, ``Quality-aware clip for blind image quality assessment,'' in \emph{Chinese Conference on Pattern Recognition and Computer Vision (PRCV)}.\hskip 1em plus 0.5em minus 0.4em\relax Springer, 2023, pp. 396--408.

\bibitem{miyata2023interpretable}
T.~Miyata, ``Interpretable image quality assessment via clip with multiple antonym-prompt pairs,'' in \emph{2023 IEEE 13th International Conference on Consumer Electronics-Berlin (ICCE-Berlin)}.\hskip 1em plus 0.5em minus 0.4em\relax IEEE, 2023, pp. 39--40.

\bibitem{zhang2023advancing}
Z.~Zhang, W.~Sun, Y.~Zhou, H.~Wu, C.~Li, X.~Min, X.~Liu, G.~Zhai, and W.~Lin, ``Advancing zero-shot digital human quality assessment through text-prompted evaluation,'' \emph{arXiv preprint arXiv:2307.02808}, 2023.

\bibitem{vaswani2017attention}
A.~Vaswani, N.~Shazeer, N.~Parmar, J.~Uszkoreit, L.~Jones, A.~N. Gomez, {\L}.~Kaiser, and I.~Polosukhin, ``Attention is all you need,'' \emph{Advances in Neural Information Processing Systems}, vol.~30, 2017.

\bibitem{tsai2007frank}
M.-F. Tsai, T.-Y. Liu, T.~Qin, H.-H. Chen, and W.-Y. Ma, ``Frank: a ranking method with fidelity loss,'' in \emph{Proceedings of the 30th Annual International ACM SIGIR Conference on Research and Development in Information Retrieval}, 2007, pp. 383--390.

\bibitem{thurstone1994law}
L.~L. Thurstone, ``A law of comparative judgment.'' \emph{Psychological Review}, vol. 101, no.~2, p. 266, 1994.

\bibitem{nichol2021glide}
A.~Nichol, P.~Dhariwal, A.~Ramesh, P.~Shyam, P.~Mishkin, B.~McGrew, I.~Sutskever, and M.~Chen, ``Glide: Towards photorealistic image generation and editing with text-guided diffusion models,'' \emph{arXiv preprint arXiv:2112.10741}, 2021.

\bibitem{rombach2022high}
R.~Rombach, A.~Blattmann, D.~Lorenz, P.~Esser, and B.~Ommer, ``High-resolution image synthesis with latent diffusion models,'' in \emph{Proceedings of the IEEE/CVF conference on computer vision and pattern recognition}, 2022, pp. 10\,684--10\,695.

\bibitem{xu2018attngan}
T.~Xu, P.~Zhang, Q.~Huang, H.~Zhang, Z.~Gan, X.~Huang, and X.~He, ``Attngan: Fine-grained text to image generation with attentional generative adversarial networks,'' in \emph{Proceedings of the IEEE Conference on Computer Vision and Pattern Recognition}, 2018, pp. 1316--1324.

\bibitem{borji2022generated}
A.~Borji, ``Generated faces in the wild: Quantitative comparison of stable diffusion, midjourney and dall-e 2,'' \emph{arXiv preprint arXiv:2210.00586}, 2022.

\bibitem{ramesh2022hierarchical}
A.~Ramesh, P.~Dhariwal, A.~Nichol, C.~Chu, and M.~Chen, ``Hierarchical text-conditional image generation with clip latents,'' \emph{arXiv preprint arXiv:2204.06125}, vol.~1, no.~2, p.~3, 2022.

\bibitem{zhou2022towards}
Y.~Zhou, R.~Zhang, C.~Chen, C.~Li, C.~Tensmeyer, T.~Yu, J.~Gu, J.~Xu, and T.~Sun, ``Towards language-free training for text-to-image generation,'' in \emph{Proceedings of the IEEE/CVF Conference on Computer Vision and Pattern Recognition}, 2022, pp. 17\,907--17\,917.

\bibitem{bao2023one}
F.~Bao, S.~Nie, K.~Xue, C.~Li, S.~Pu, Y.~Wang, G.~Yue, Y.~Cao, H.~Su, and J.~Zhu, ``One transformer fits all distributions in multi-modal diffusion at scale,'' \emph{arXiv preprint arXiv:2303.06555}, 2023.

\bibitem{zhang2023adding}
L.~Zhang, A.~Rao, and M.~Agrawala, ``Adding conditional control to text-to-image diffusion models,'' in \emph{Proceedings of the IEEE/CVF International Conference on Computer Vision}, 2023, pp. 3836--3847.

\bibitem{antkowiak2000final}
J.~Antkowiak, T.~J. Baina, F.~V. Baroncini, N.~Chateau, F.~FranceTelecom, A.~C.~F. Pessoa, F.~S. Colonnese, I.~L. Contin, J.~Caviedes, and F.~Philips, ``Final report from the video quality experts group on the validation of objective models of video quality assessment march 2000,'' \emph{Final report from the video quality experts group on the validation of objective models of video quality assessment march 2000}, 2000.

\bibitem{dosovitskiy2020image}
A.~Dosovitskiy, L.~Beyer, A.~Kolesnikov, D.~Weissenborn, X.~Zhai, T.~Unterthiner, M.~Dehghani, M.~Minderer, G.~Heigold, S.~Gelly \emph{et~al.}, ``An image is worth 16x16 words: Transformers for image recognition at scale,'' \emph{arXiv preprint arXiv:2010.11929}, 2020.

\bibitem{zhang2018blind}
W.~Zhang, K.~Ma, J.~Yan, D.~Deng, and Z.~Wang, ``Blind image quality assessment using a deep bilinear convolutional neural network,'' \emph{IEEE Transactions on Circuits and Systems for Video Technology}, vol.~30, no.~1, pp. 36--47, 2018.

\bibitem{saha2023re}
A.~Saha, S.~Mishra, and A.~C. Bovik, ``Re-iqa: Unsupervised learning for image quality assessment in the wild,'' in \emph{Proceedings of the IEEE/CVF Conference on Computer Vision and Pattern Recognition}, 2023, pp. 5846--5855.

\bibitem{zhou2021lafite}
Y.~Zhou, R.~Zhang, C.~Chen, C.~Li, C.~Tensmeyer, T.~Yu, J.~Gu, J.~Xu, and T.~Sun, ``Lafite: Towards language-free training for text-to-image generation. arxiv 2021,'' \emph{arXiv preprint arXiv:2111.13792}, vol.~2, 2021.

\end{thebibliography}

\end{document}